\documentclass[nopreprintline,12pt]{elsarticle}

\usepackage{float} 
\usepackage{amssymb}
\usepackage{amsmath}
\usepackage{gensymb}
\journal{Robotics and Autonomous Systems}

\begin{document}

\begin{frontmatter}



\title{Temperature Driven Multi-modal/Single-actuated Soft Finger}

\affiliation[label1]{organization={Graduate School of Engineering Science, Osaka University},%Department and Organization
            addressline={Machikaneyamacho}, 
            city={Toyonaka},
            postcode={5600043}, 
            state={Osaka},
            country={Japan}}

\affiliation[label2]{organization={ICPS Research Center, National Institute of Advanced Industrial Science and Technology (AIST)},
             addressline={2-4-7 Aomi},
             city={Koto-ku},
             postcode={1350064},
             state={Tokyo},
             country={Japan}}

\author[label1]{Prashant Kumar} %% Author name
\author[label1]{Weiwei Wan}
\author[label1,label2]{Kensuke Harada}

%% Abstract
\begin{abstract}
Soft pneumatic fingers are of great research interest. However, their significant potential is limited as
most of them can generate only one motion, mostly bending. The conventional design of soft fingers
doesn’t allow them to switch to another motion mode. In this paper, we developed a novel multi-modal
and single-actuated soft finger where its motion mode is switched by changing the finger’s temperature.
Our soft finger is capable of switching between three distinctive motion modes: bending, twisting, and
extension-in approximately five seconds. We carried out a detailed experimental study of the soft finger and
evaluated its repeatability and range of motion. It exhibited repeatability of around one millimeter and
a fifty percent larger range of motion than a standard bending actuator. We developed an analytical
model for a fiber-reinforced soft actuator for twisting motion. This helped us relate the input pressure to
the output twist radius of the twisting motion. This model was validated by experimental verification.
Further, a soft robotic gripper with multiple grasp modes was developed using three actuators. This
gripper can adapt to and grasp objects of a large range of size, shape, and stiffness. We showcased its
grasping capabilities by successfully grasping a small berry, a large roll, and a delicate tofu cube.
\end{abstract}



%% Keywords

\end{frontmatter}



\section{Introduction}
\label{sec1}
%% Labels are used to cross-reference an item using \ref command.
Soft actuators have emerged as a vital component in robotic applications requiring gentle handling, intricate manoeuvres, high flexibility, and adaptive interaction with the environment\cite{hughes2016soft,li2022soft,el2020soft,kim2019review}. However, most existing soft actuators are limited to executing only a single type of motion, such as bending, twisting, or extension\cite{yao2024multimodal,shintake2018soft,haines2014artificial,feng2020controlled}. This limitation significantly restricts their versatility for complex tasks that demand multi-directional and adaptive motion.  A wide range of soft robotics applications, from manipulators to locomotion systems, could benefit from actuators with multi-motion capabilities\cite{patel2023highly,chi2022bistable,chen2020soft}. Such actuators enable access to a broader range of positions in 3D space, enhancing grasping options and expanding the scope of adaptable and dexterous movements. However, achieving this level of functionality has remained a significant challenge, as most designs either require complex control systems with multiple inputs\cite{balak,Liao,guan2020novel} or rely on rigid components\cite{Jinag_scaffold}, compromising the inherent softness and adaptability of these systems.

Among soft actuators, Fluidic Elastomer Actuators(FEAs) have been a popular choice due to their safe operation and high energy efficieny\cite{al2016power,gonzalez2023soft}. FEAs such as PneuNets\cite{polygerinos_roboglove,shintake2018soft,mosadegh2014pneumatic}, utilize elastomeric materials with embedded air channels that expand asymmetrically under pressure to produce bending motions. Similarly, fiber-reinforced actuators\cite{bishop2012design,galloway2013mechanically} guide motion into controlled bending, twisting, or extension by constraining expansion with embedded fibers. Despite these advancements, such designs remain inherently constrained in motion diversity, capable of performing only a single, straightforward motion—bending—thereby limiting their applicability across a broader range of tasks\cite{Polygerinos_modelling}. Efforts to enhance motion diversity have included the use of bi-directional bending actuators and multi-stimulus designs. For instance, Bilodeau et al.\cite{bilodeau2018design} achieved bi-directional bending by assembling two antagonistic PneuNet actuators, while Yang et al.\cite{yang2017sma} combined FEAs with Shape Memory Polymers (SMPs) to enable bending at different points along the actuator. However, the reliance on high-temperature stimuli (up to 50°C and more) in SMP-based designs limits their suitability for handling temperature-sensitive or fragile objects. Similarly, magnetic stimulus-based actuators, such as the bi-stable designs proposed by Dai et al.\cite{dai2022bionic}, achieved rapid state switching but were confined to 2D planar motion. This limited range makes them unsuitable for applications requiring dexterous movements in a 3D space.\\
\begin{figure}[t]
%\centering
\includegraphics[width=13.4cm]{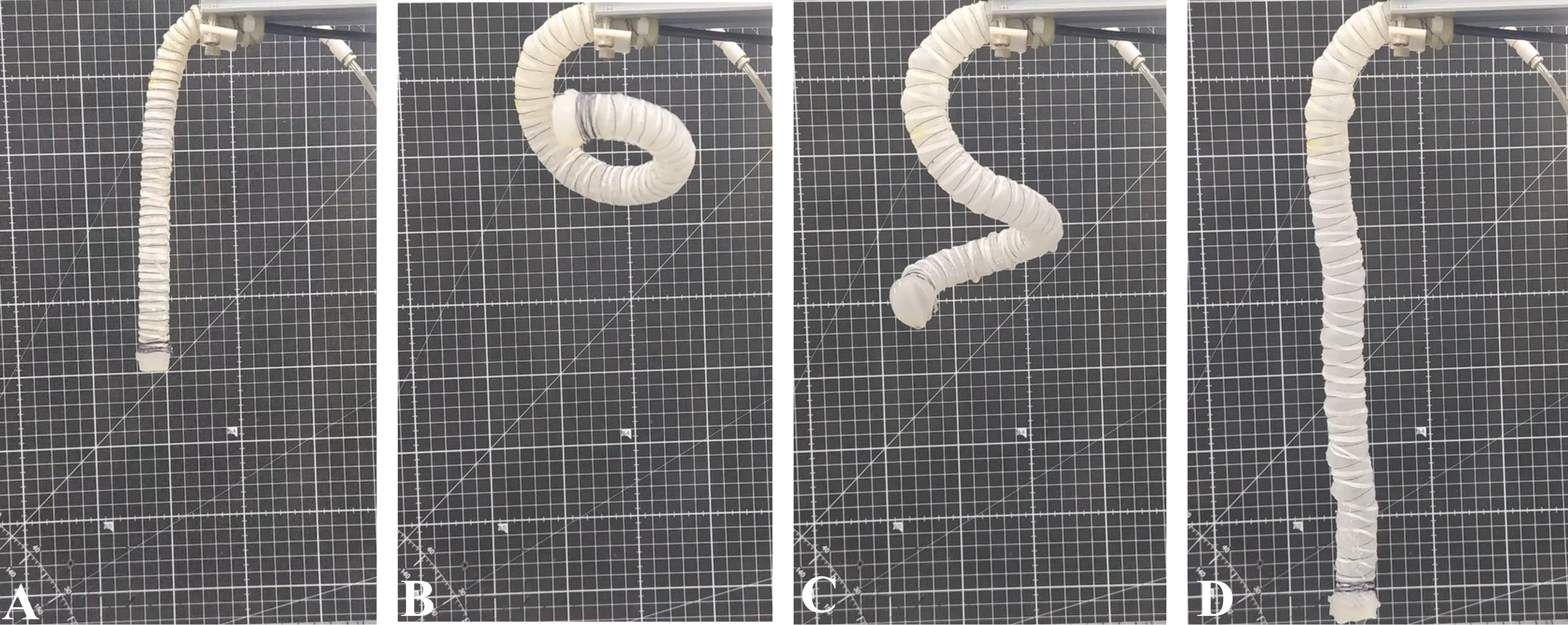}
\caption{Developed actuator and it's three motion modes (A) Actuator in resting state. (B) Bending mode. (C) Twisting mode. (D) Extension mode.}\label{FigureLabel1}
%\end{SCfigure*}
\end{figure}%[tbhp]
Some researchers have explored helical or coiled designs to introduce 3D motion capabilities. Martinez et al.\cite{Martinez} and Yuan et al.\cite{yuan2020design} developed FEAs capable of coiling to achieve 3D motions, while Gu et al.\cite{Gu} and Wang et al.\cite{Wang_twist} utilized a Pneunet-based helical motion actuator to grasp larger objects. Despite their ability to move in 3D, these actuators are still restricted to a single motion type, such as coiling or twisting, limiting their versatility.\\
There were further attempts to create actuators capable of two motion modes. Jiang et al.\cite{Jinag_scaffold} introduced a scaffold-reinforced actuator capable of bending and twisting by altering the scaffold configuration. However, the inclusion of rigid elements compromised the softness of the actuator, reducing its ability to interact safely with delicate objects. Another popular approach has been coupling multiple actuators in parallel or series to achieve motion diversity\cite{connolly2017automatic,Liao,Huang,Kan,huang2020multimodal}. Abondance et al. \cite{Abondance} connected two actuators in parallel to produce bending and yaw motions, while Balak et al.\cite{balak2020multi} used three bellow actuators to generate three distinct motions. Similarly, Liao et al. \cite{Liao} combined an extending actuator with two twisting actuators to create a snake-like climbing robot. However, these approaches require a high number of inputs and precise synchronization, significantly increasing the complexity of control systems.\\

In response to these limitations, we propose a novel soft actuator capable of performing and switching between three distinct motions—bending, twisting, and extension—using only two simple inputs: a single pressure input and a temperature stimulus. By integrating a soft, paper-thin variable-stiffness material (Humofit\cite{Humofit}) into the actuator, we achieve quick and seamless motion switching while preserving flexibility. Unlike existing solutions, our design eliminates the need for additional actuators or rigid components, preserving the system's inherent softness and adaptability while ensuring a straightforward and efficient control approach. The ability to execute and seamlessly switch between three distinct motions expands the actuator's potential for diverse robotic applications. To demonstrate its capabilities, we implemented our actuator in a robotic gripper application. Using three identical actuators, the gripper utilizes the combination of bending, twisting, and extension to achieve various grasping modes. These modes, combined with the deformable material of the actuators, allow for successful grasping objects of varying size, shape, and stiffness, highlighting the versatility and practicality of the developed actuator. Furthermore, we characterized the actuator’s motion range and repeatability and developed an analytical model to establish the relationship between input pressure and output twist radius, which was experimentally validated.

\noindent The main contributions are:\\
1.  Development of a single-chamber soft actuator capable of three distinct motions: bending, twisting, and extension.\\
2.  Integration of a variable-stiffness material with fiber reinforcement to enable motion mode switching via thermal stimulus.\\
3.  Analytical modeling to establish and experimentally validate the relationship between input pressure and output twist radius for twisting motion.\\
4.  Demonstration of the actuator’s versatility through a robotic gripper capable of adaptive grasping for objects with varying size, shape, and stiffness.\\

This actuator’s simplicity in control, versatility in motion diversity, and speed in motion mode change makes it a significant advancement in soft robotics, paving the way for more adaptable, compact, and efficient multimotion soft robotic systems.

%\begin{SCfigure*}[\sidecaptionrelwidth][t!]
\begin{figure*}[t]
\centering
\includegraphics[width=13.4cm]{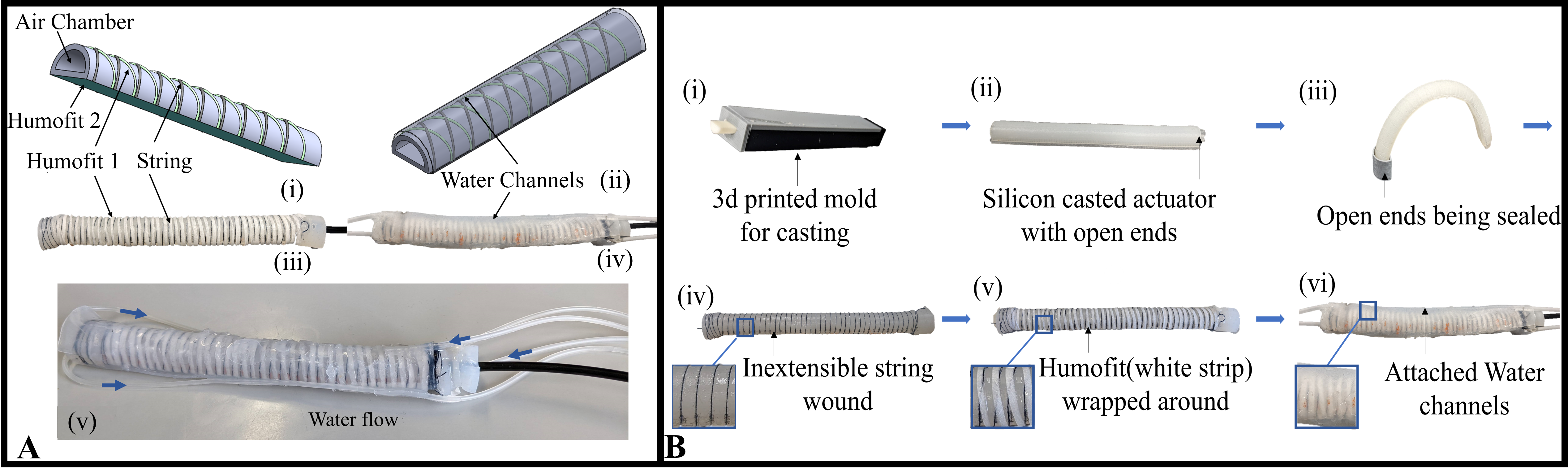}
\caption{Structure and construction of the actuator (A) (i)Inner components of the soft actuator (CAD image). (ii) Outer components of the soft actuator. (iii)Inner structure of the soft actuator. (iv) Outer structure of the soft actuator with water channels i.e. A complete actuator. (v) Flow
of water used to heat/cool humofit. (B) Actuator’s construction process. (i)Silicon casting. (ii) Molded Silicon body. (iii) Sealing off open ends. (iv) Winding non-extensible thread. (v) Winding Humofit. (vi) Attaching water channels over the
actuator.
}\label{FigureLable2}
%\end{SCfigure*}
\end{figure*}%[tbhp]

\section{Soft Actuator}
In this section, we explain the details of the actuator's structure, its construction process, and the working principle behind the multi-motion ability of the actuator.

\subsection{Structure}
The developed soft actuator’s structure falls in the fiber-reinforced soft actuator category\cite{polygerinos_roboglove}. The main difference from a standard structure is the restraining elements being made of Humofit\cite{Humofit} (a product manufactured by Mitsui Chemicals Group in Japan). Humofit is a variable stiffness material that uses temperature change as its stimulus. This material achieves a stiff state when subjected to cold temperatures i.e. 10 \degree C and below. It achieves a stretchable state when its temperature is increased i.e. 28 \degree C and higher. The change in material’s temperature happens within a few seconds hence our actuator can also change its modes within the same amount of time. Temperature-based stress-strain curve and other physical properties can be found on Humofit’s website\cite{Humofit}.

The soft actuator consists of five major parts. The first part is a core silicon actuator(Fig.\ref{FigureLable2}A). The actuator has two fibers acting as reinforcing elements. One is a 2mm thin humofit thread denoted as Humofit1(Fig.\ref{FigureLable2}A(i)) while the other is an in-extensible nylon string. The fourth component is a 20 mm wide humofit strip placed under the silicon core as its passive layer. The final component is an outer silicon covering denoted as water channels(Fig.\ref{FigureLable2}A(ii,iv)). It allows hot/cold water to be passed over humofit elements during operation.

The soft actuator has a semi-circular cross section. It is 170 mm in length, 12 mm in radius, and 3mm in wall thickness. The pitch and angle of the fiber windings are 4mm and 5\degree respectively. These parameters were chosen to enhance the efficiency of output motion for an input pressure. These values were decided with the help of \cite{connolly2017automatic,Wang_structure,Connolly_fangle} which has explained the role of these parameters in an actuator’s performance in detail.
%\begin{SCfigure*}[\sidecaptionrelwidth][t!]
\begin{figure*}[t]
\centering
\includegraphics[width=11.4cm]{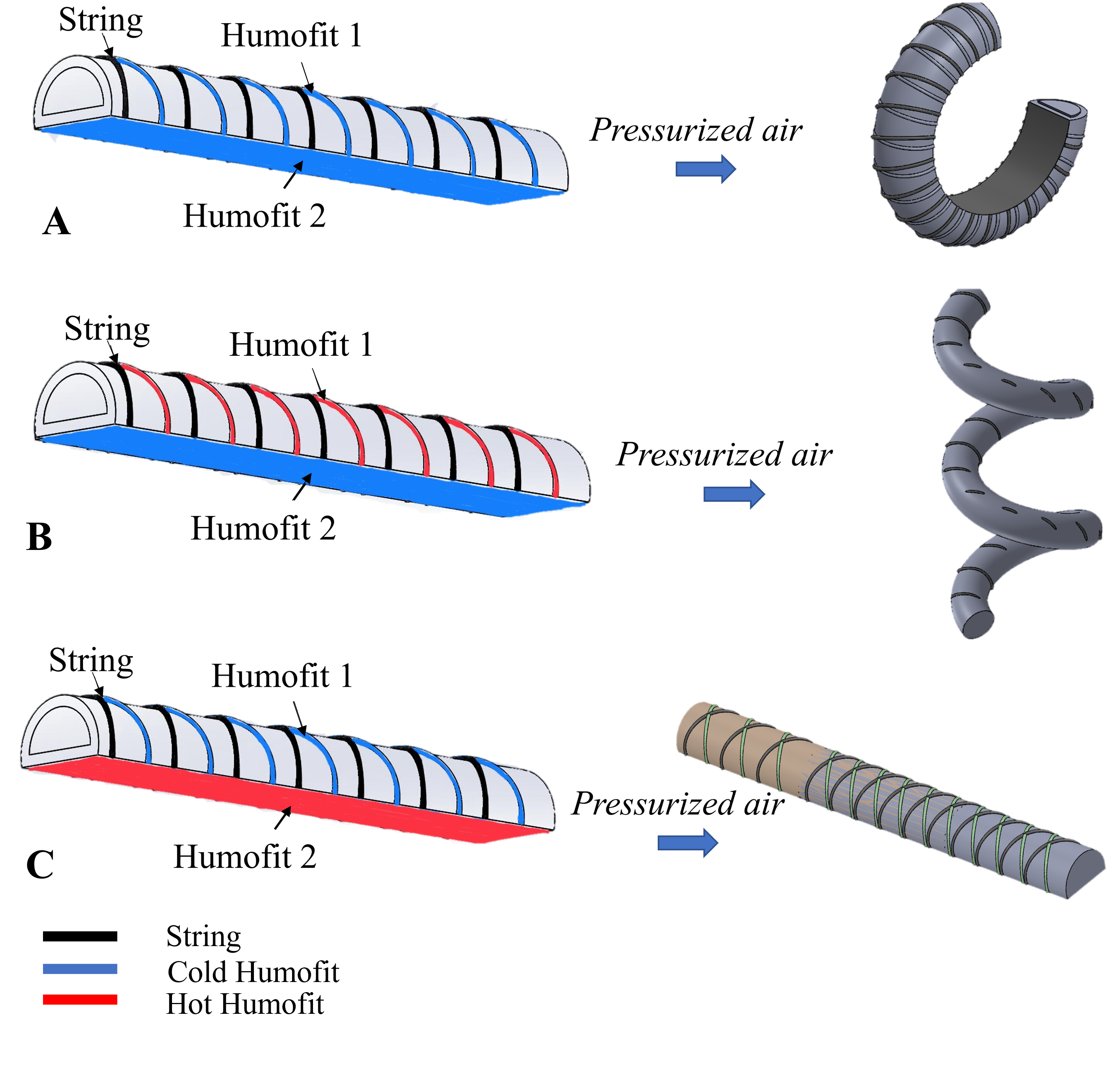}
\caption{Working principle of the actuator (A)Bending Mode: String, Humofit1, and Humofit2 are stiff as they are cold(blue colour). (B)Twisting Mode: Humofit1 is hot hence stretchable (red colour) while the string and Humofit2 are cold hence stiff(blue colour). (C) Extension
Mode: Humofit1 and string are stiff while Humofit2 is hot (red).
}\label{FigureLabel3}
%\end{SCfigure*}
\end{figure*}%[tbhp]

\subsection{Working principle}
%The actuator has two fibers acting as reinforcing elements. (shown in Fig.\ref{FigureLabel2}A ). A string is used as the first fiber and a 1mm wide strip of Humofit is used as the other fiber denoted as Humofit 1. The actuator also has a passive layer made of Humofit at the bottom. This is denoted as Humofit 2.
Humofit is a material capable of varying its stiffness as a function of its temperature. The material becomes extremely rigid and inextensible at temperatures below 10\degree C. On the other hand, it shows high stretchability and least resistance to forces at temperatures over 40\degree C\cite{Humofit}. \\

To stimulate Humofit, cold or hot water is passed from a 5V DC water pump through thin silicon tubes to the water channels(Fig.\ref{FigureLable2}A(ii),(iv),(v)). By changing the temperature of Humofit elements, the actuator can switch between three different configurations which results in three different motions (Fig.\ref{FigureLabel3}). To generate bending, no temperature changes are needed. Bending is the default motion. When pressure is applied to the actuator, both reinforcing fibers (String and Humofit1) apply similar forces leading to a bending motion(Fig.\ref{FigureLabel3}A). To generate twisting motion, hot water is passed over Humofit 1. The hot water raises the stretchability of Humofit 1 to such a high extent that its constraining force on the actuator becomes negligible. This creates a new configuration of reinforcing elements that favors twisting
motion(Fig.\ref{FigureLabel3}B). Similarly, to achieve extension motion, hot water is passed over Humofit 2 which makes its constraining ability ineffective, leading to a new configuration that favours extension motion(Fig. \ref{FigureLabel3}C). It was observed that the actuator could switch between two motions in approximately five seconds(Movie S3) and it maintains the stimulated stiffness for at least one minute.

%\begin{SCfigure*}[\sidecaptionrelwidth][t!]
\begin{figure}[]
\centering
\includegraphics[width=11.4cm]{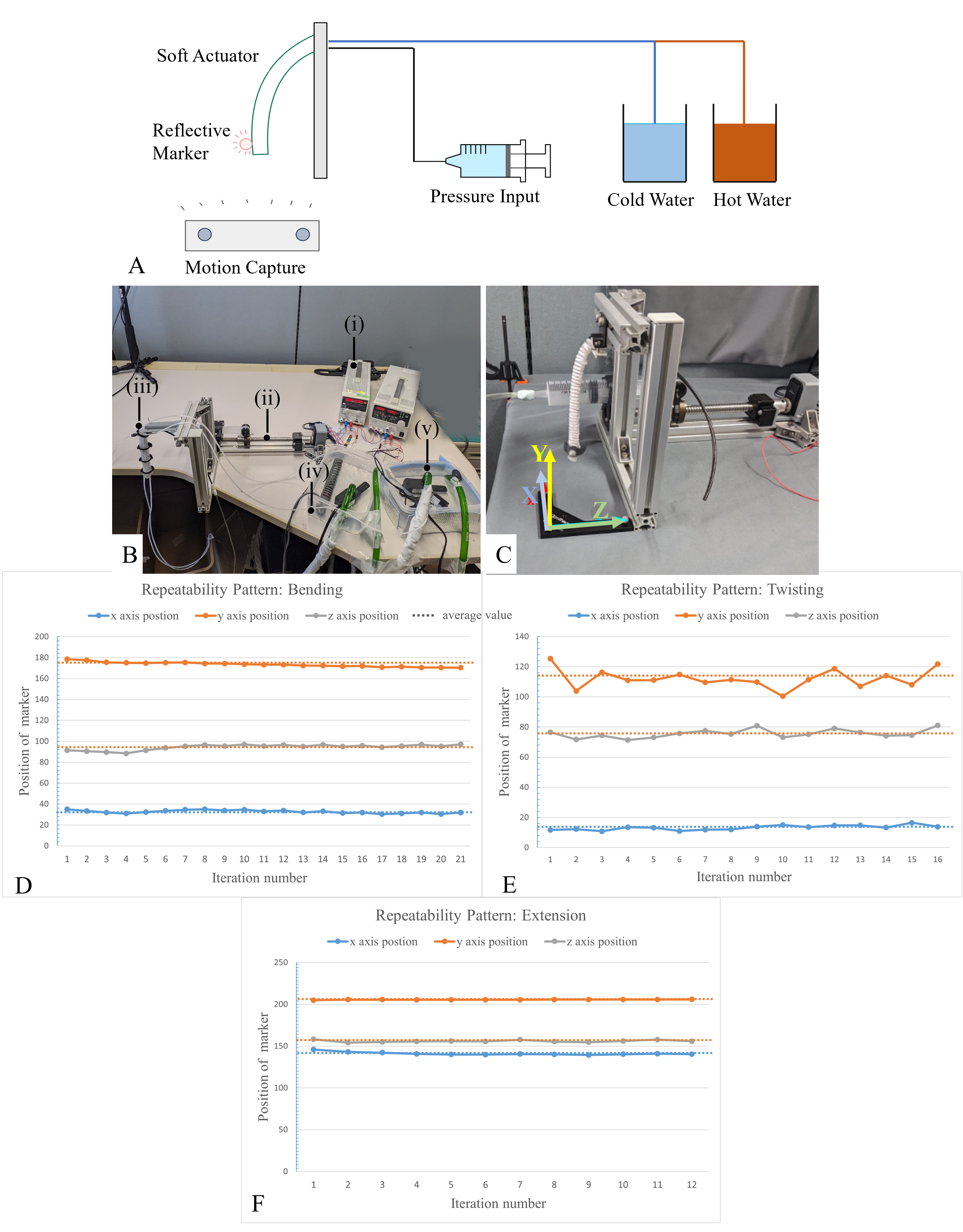}
\caption{Experimental setup and output of the actuator's repeatability test (A) Main components of the experiment. (B) (i) Power supply. (ii) Pressure input. (iii) Soft actuator. (iv) Hot water. (v) Cold water. (C) Origin and axes for measurement. (D) Bending Repeatability output. (E) Twisting Repeatability output. (F) Extension Repeatability output.
}\label{FigureLabel4}
%\end{SCfigure*}
\end{figure}%[tbhp]

\subsection{Construction}
 The soft actuator was constructed by silicon cast molding, following the process described in\cite{Polygerinos_modelling,polygerinos_roboglove} with certain additions. Ecoflex-0030 was chosen as the construction material due to its stretchable nature. The first step is to create a mold according to the actuator's planned height, width, and wall thickness. The mold was created with z-ultrat material in a 3D printer. Firstly, silicon is poured into the mold and allowed to rest for four hours(Fig.\ref{FigureLable2}B(i),(ii)). The next step is to seal the ends of the actuator with silicon. There are several ways to do it\cite{Polygerinos_modelling}. In this paper, the ends were sealed using silicon to ensure softness throughout the actuator and to prevent any possible air leaks(Fig.\ref{FigureLable2}B(iii)). This is followed by adding reinforcing elements to the actuator. A humofit strip whose width equals the actuator's width is placed at the bottom of the actuator(Humofit2). A nylon thread is wound around the actuator, followed by applying a small amount of silicon over the thread(Fig.\ref{FigureLable2}B(iv)). This ensures that the thread stays in the same position during the operation. The next step is to wind a Humofit thread(Humofit1) which is made by cutting thin slices of humofit and sticking them together with heating equipment(Fig.\ref{FigureLable2}B(v)). A thin layer of silicon is applied over the actuator again to ensure that both fibers don't get displaced during the expansion of the silicon body. The last step is to pierce a small hole and put in a pneumatic cable for air input(Fig.\ref{FigureLable2}B(vi)). 

 Since the soft actuator is being constructed manually, there are possibilities of error at multiple stages. Certain steps were taken to ensure as minimum error as possible. The motion output of the actuator depends on the fiber winding's pitch. To ensure uniform winding, small indents were made on the sides of the soft actuator called guides. These act as guides along which a user can uniformly wind the fibers. Also, preventing air bubble formation during the casting process will ensure a durable actuator.

\section{Characteristics}
As a newly developed actuator, it is important to evaluate its capabilities. This allows us to quantify the advantages of the actuator. We evaluated its performance along two separate parameters:

\subsection{Repeatability}

 The first evaluated parameter was Repeatability. The goal of the repeatability experiment was to evaluate whether the actuator is fit for precision-based tasks or not. The setup used in the repeatability experiment can be seen in Fig. \ref{FigureLabel4} B, C.

In the experiment, the actuator was actuated 30 times for each motion mode: bending, twisting and extension. Each actuation cycle lasting 13 seconds. A reflective marker of a diameter 5mm was attached at the end of the soft actuator(Fig. \ref{FigureLabel4}C). The motion data of this marker was captured using the Optitrack motion capture system. To maintain the temperature, hence the stiffness of humofit, the desired temperature water was passed over the actuator via water channels(Fig. \ref{FigureLable2}A). An aquarium temperature regulating setup was used to store hot/cold water and maintain its temperature during the experiment.
The average deviation in repeatability values for bending, twisting and extension motion were 2.1 mm, 3.29 mm, and 1.1 mm respectively. Twisting, being a 3-dimensional motion, was subjected to more error, hence a bigger deviation. The average repeatability deviation of the actuator was found to be 2.16mm. 
%\begin{SCfigure*}[\sidecaptionrelwidth][t!]
\begin{figure*}[tbhp]
\centering
\includegraphics[width=11.4cm]{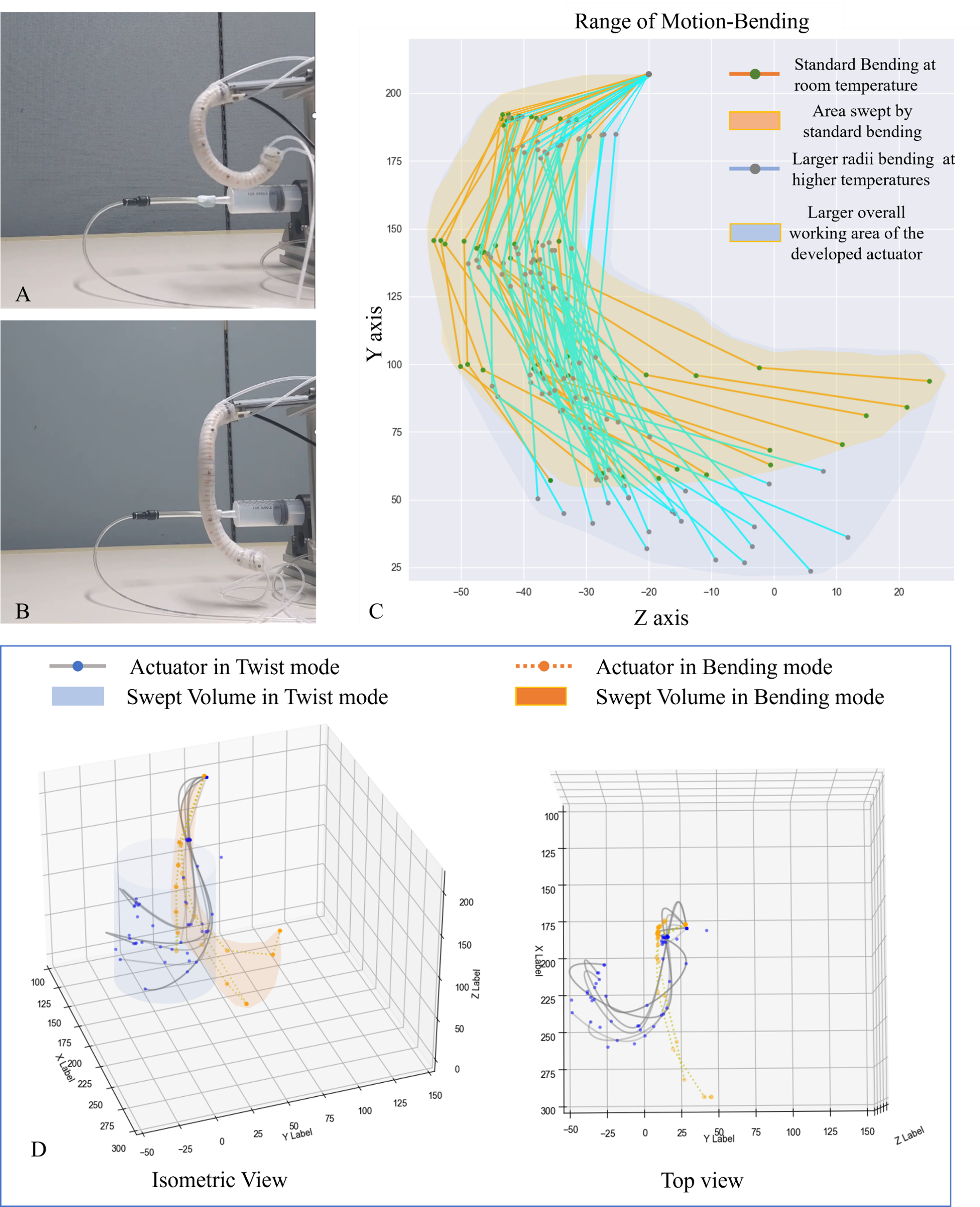}
\caption{Large range of motion exhibited by the developed actuator (A) Bending at 25\degree C, 20kPa. (B) Bending at 45\degree C, 20kPa. (C) Sweeping area of the bending actuator. (D) Volume swept by the actuator in twist mode compared to the standard bending mode presented via two views.}\label{FigureLabel5}
%\end{SCfigure*}
\end{figure*}%[tbhp]

\subsection{Range of Motion}
This experiment evaluated the actuator’s overall range of motion. The actuator was provided with the same input pressure while the temperature of Humofit2 was changed at every actuation. The temperature of Humofit1 was kept at 25\degree C (Room Temperature) while the temperature of Humofit2 was varied by 10 degrees, starting from 5\degree C to 55\degree C. Four reflective markers were placed on the actuator. These markers were traced by the OptiTrack motion capture system which helped us obtain the sweeping volume of the actuator. The developed actuator performs a standard bending motion at room temperature and lower (0\degree C - 25\degree C). The volume swept during this standard bending motion is shown in orange (Fig. \ref{FigureLabel5}C). As the temperature is increased, the actuator tends to exhibit a combination of extension and bending motion, resulting in a larger bending radius(Fig. \ref{FigureLabel5}A, B). The actuator at 35,45 and 55\degree C is represented by blue lines and the increased sweeping volume after including the motion at 35,45 and 55\degree C is shown as the blue region. 
As compared to the volume swept by the actuator with its standard unstimulated bending mode (represented by orange in Fig. \ref{FigureLabel5}), the actuator can sweep roughly 50 percent more volume when the bending-extension mode is utilized as well(blue zone in Fig. \ref{FigureLabel5}). Using the convex hull method, we obtained the volume swept by the bending motion and pure extension to be 10035 mm$^3$ and 5019mm$^3$ respectively. 

Similarly, we performed a range of motion evaluation for the twisting mode of the actuator as shown in Fig. \ref{FigureLabel5}D. The coordinate system used is similar to the one shown in Fig. \ref{FigureLabel7}C. The orange and the blue regions represent the volume swept in the standard bending mode and the twisting mode respectively. We used the convex hull method and found out that in the twist mode, the actuator sweeps through approximately 305367 mm$^3$ of volume, apart from the existing working region of the actuator. Twisting being a three-dimensional motion allows the actuator to access a large zone around the bending region of the actuator. Overall swept volume of the actuator is 320421 mm$^3$. Considering this increased range of motion in all three instances, it is safe to conclude that the developed multimodal actuator has a significantly larger range of motion than a standard bending actuator.

%\begin{SCfigure*}[\sidecaptionrelwidth][t!]
\begin{figure}[tbhp]
\centering
\includegraphics[width=13.4cm]{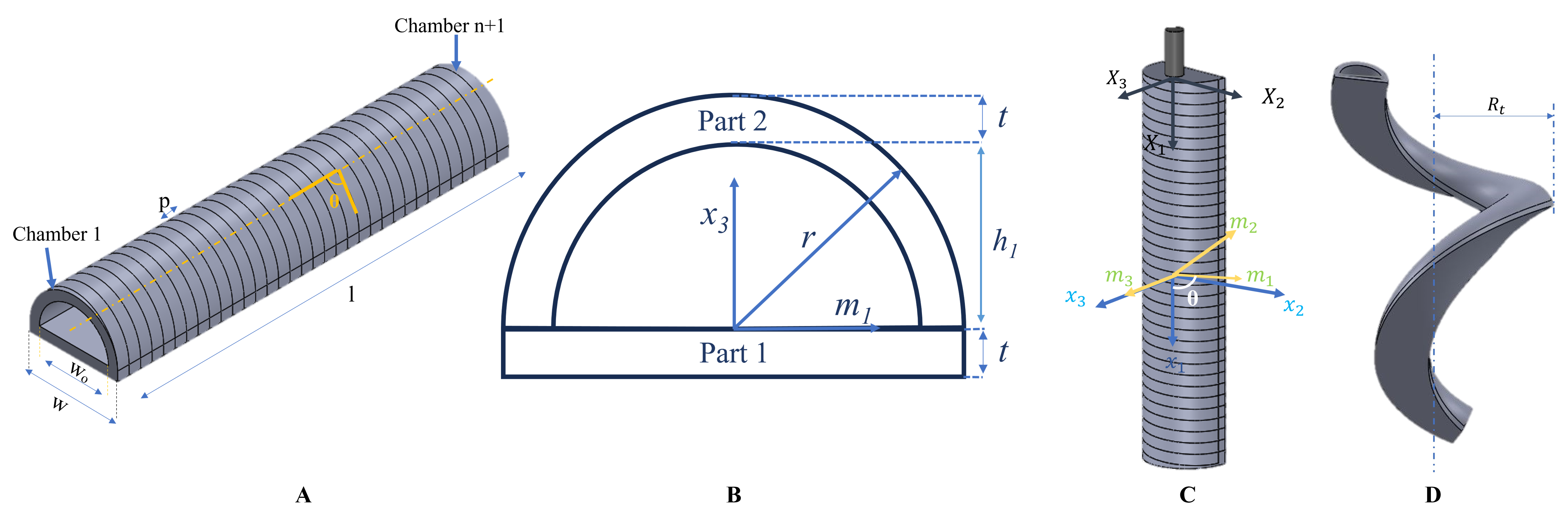}
\caption{(A) Actuator parameters. (B) Cross-section of the actuator. (C) Coordinate systems for twist output. (D) Twist Radius.}\label{FigureLabel6}
%\end{SCfigure*}
\end{figure}%[tbhp]

\section{Analytical Modelling}
It is important to establish a relationship between the input pressure and the output motion of an actuator. The analytical model for Bending and extension motion for a fiber reinforced actuator can be found here\cite{polygerinos_roboglove,connolly2017automatic}. 
%In addition to the model for bending motion,\cite{polygerinos_roboglove}
We developed a Pressure-Twist radius relation for a twisting actuator. Further, we verified the correctness of the developed model via motion capture experiments.

\subsection{Theoretical Relation}

%The modeling for twist motion is based on the work of Gu \textit{et al.} \cite{Gu}, who developed a Pressure- Twist radius relation for a Square cross-section, Pneunet actuator. We have modified this to achieve the relation between Input pressure P and Twist Radius $R_t$ for a semi-circular cross-section, fiber-reinforced actuator.
The modeling approach is based on the minimum potential energy method and the continuum rod theory. This method takes into consideration the geometric parameters and the material elasticity of the actuator. So far, this approach has been used to model twist motion for a square cross-section pneunet actuator\cite{Gu}. In this section, we present a relation between Input pressure P and Twist Radius $R_t$ for a semi-circular cross-section, fiber-reinforced actuator.

The equilibrium state is obtained by using the minimization of Potential energy $ \prod $. We treated the region between two fiber windings as individual chambers. For a Fiber Reinforced actuator with width $w$, pitch $p$, angle of winding $\theta$ and number of windings $n$:\\
Total Potential Energy = $\prod$  \\

$$\prod = (n+1)( W_m - W_F) $$      \\
Where, \\
 $W_m$ = Elastic energy stored in the silicon elastomer \\
      $W_F$ = Work done by the supplied pressure in each chamber\\
      $n+1$ = Total number of chambers formed \\

$$W_m = W_{m1} + W_{m2} $$ \\
Where $W_{m1}$  and  $W_{m2} $ are the elastic energy stored in Part 1 and Part 2, as shown in Fig. \ref{FigureLabel6}B. 
Also, the strain energy density function per unit volume\cite{holzapfel2002nonlinear} is given by 
\begin{equation}
    \label{eq1}
W_s = \frac{1}{2}\frac{E}{1+v}J_2 + \frac{Ev(1-2v)}{1+v}J_1^2
\end{equation}
Where, \\
$J_1$= First Strain Invariant\\
$J_2$= Second Strain Invariant\\
E= Young Modulus\\
v= Poisson's Ratio\\

The elastic strain energy in Part 1 can be calculated by: 
\begin{equation}
    \label{eq2}
    W_{m1} = p\frac{W}{cos\theta }\int_{x_3=0}^{t} W_s \,dx_3
\end{equation}
 \\
where \textit{t} is the wall thickness of the actuator. Now, to evaluate the Strain energy in part 2, we consider an element w\textsuperscript{'} which is the width in a quadrant at height $x_3$. The elastic strain energy of part2 will be given by:

%%$$ W_{m2} = p\int W_s\,dA$$
$$W_{m2} = p\int W_s\ 2w^{'} \,dx_3 $$
$$W_{m2} = p\int_{r-t}^{r} W_s\ 2w^{'}\,dx_3$$
\begin{equation}
    \label{eq3}
W_{m2} = p\int_{r-t}^{r} W_s\ 2\sqrt{r^2 - x_{3}^2}\,dx_3 
\end{equation}
Since, $(w^{'})^{2} + x_{3}^2 = r^2  $ (circle's property)\\ 

The coordinate systems used here are the global coordinate $X_1$-$X_2$-$X_3$, local coordinate $x_1$-$x_2$-$x_3$ and chamber orientation coordinate $m_1$-$m_2$-$m_3$(Fig. \ref{FigureLabel6}C).
$e_{11}$,$e_{22}$ and $e_{33}$ being the principle components at $x_3$=0. $k_1$,$k_2$and $k_3$ being the bending curvatures and $\epsilon$ being the strain\cite{Gu}.
Now, The involved parameters can be expressed by using Euler- Bernoulli beam theory as:
$$J_1 = \epsilon_{11}^{(r)} + \epsilon_{22}^{(r)} + \epsilon_{33}^{(r)} $$
$$J_2 = (\epsilon_{11}^{(r)})^2 + (\epsilon_{22}^{(r)})^2 + (\epsilon_{33}^{(r)})^2$$
$$\epsilon_{11}^{(r)}=e_{11} + x_{3}\kappa_{1}$$
$$\epsilon_{22}^{(r)}=e_{22} + x_{3}\kappa_{2}$$
$$\epsilon_{33}^{(r)}=e_{33} + x_{3}\kappa_{3}$$

\noindent Now, Work done by the supplied pressure in each chamber = $W_F$

$$W_F= P \Delta V $$
The change in volume $\Delta V$ is given by $c  h_1  \frac{w}{sin\theta}  p  \epsilon_{22}^m$ where $h_1$ and $\frac{w}{sin\theta}$ are the height and width of the chamber and the length change in the $x_1$ direction is given by $p \epsilon_{22}^m$. Hence, $W_F$ can be expressed as:
\begin{equation}
W_F= P  c  h_1  \frac{w}{sin\theta}  p  \epsilon_{22}^m     \label{eq4}
\end{equation}
\begin{equation} \label{eq5}
W_F= c P h_1 \frac{w}{sin\theta} p [\epsilon_{11}^r \sin^2{(\theta - \phi)}  + \epsilon_{22}^r \cos^2{(\theta - \phi)}] 
\end{equation}
\\
Based on the obtained $W_{m}$ and $W_{F}$, the total potential energy $\prod$ can be expressed as a function of seven independent variables where $q$ is the gradient of strain component along $x_3$ axis and $\phi$ is the phase angle\cite{Gu}. Since the changes in these variables are minimum at the state of equilibrium, we can say that: 

\begin{equation}
\label{eq6}
%\centering
\resizebox{0.85\hsize}{!}{$
    \frac{\partial\prod}{\partial k_1} =0,
    \frac{\partial\prod}{\partial k_2} = 0,\frac{\partial\prod}{\partial \phi} = 0,
    \frac{\partial\prod}{\partial e_{11}} = 0,\frac{\partial\prod}{\partial e_{22}} = 0,   
    \frac{\partial\prod}{\partial e_{33}} = 0,
    \frac{\partial\prod}{\partial q} =0$}
\end{equation}
%\centering
%\resizebox{0.45\hsize}{!}{$
%    \frac{\partial\prod}{\partial e_{22}} = 0,   
%    \frac{\partial\prod}{\partial e_{33}} = 0,
%    \frac{\partial\prod}{\partial q} =0$}

Converting the above relation from the principle strain coordinates to global coordinates $X_{1}-X_{2}-X_{3}$, we can obtain the Radius of twist $R_t$. \\
\begin{equation}
\label{eq7}
    R_t = \frac{\kappa_1 + \kappa_2 + (\kappa_1 - \kappa_2)\cos{2\phi}}{(\kappa_1)^2 + (\kappa_2)^2 + (\kappa_1 - \kappa_2)(\kappa_1 + \kappa_2)\cos{2\phi}}
\end{equation}

%To use this relation, enter the obtained values from Equation (1) (2) (3)to obtain the values of parameters. This further once substituted in equation () will provide the relation between the Radius of twist Rt and input pressure P. We have shared our relation and the corresponding parameter values here. Also, a correction factor cf is used to fit the results closer to the experimental values. \\

To use these relations for an actuator, first obtain $W_{m1}$, $W_{m2}$, and $W_F$ from Eq.\ref{eq2},\ref{eq3}, and \ref{eq4}. Solve Eq.6 using a mathematical software/tool to obtain the values of the constants involved. Finally, substitute these constant values into the equation of Rt shown in Eq.\ref{eq7}. This will generate a relationship between twist radius Rt and input pressure P which can be used for twist radius prediction.
We chose Young Modulus E= 125, Poissons ratio v=0.5 for silicon according to\cite{Gu,Liao}. We used a correction factor to bring the predicted results closer to the observed results i.e. to make the model more suitable to a specific actuator. We chose the correction factor c as 0.003(Eq. \ref{eq5}). This was obtained from\cite{Gu} and after trial and error. \\

\begin{figure*}[t]
\centering
\includegraphics[width=13.4cm]{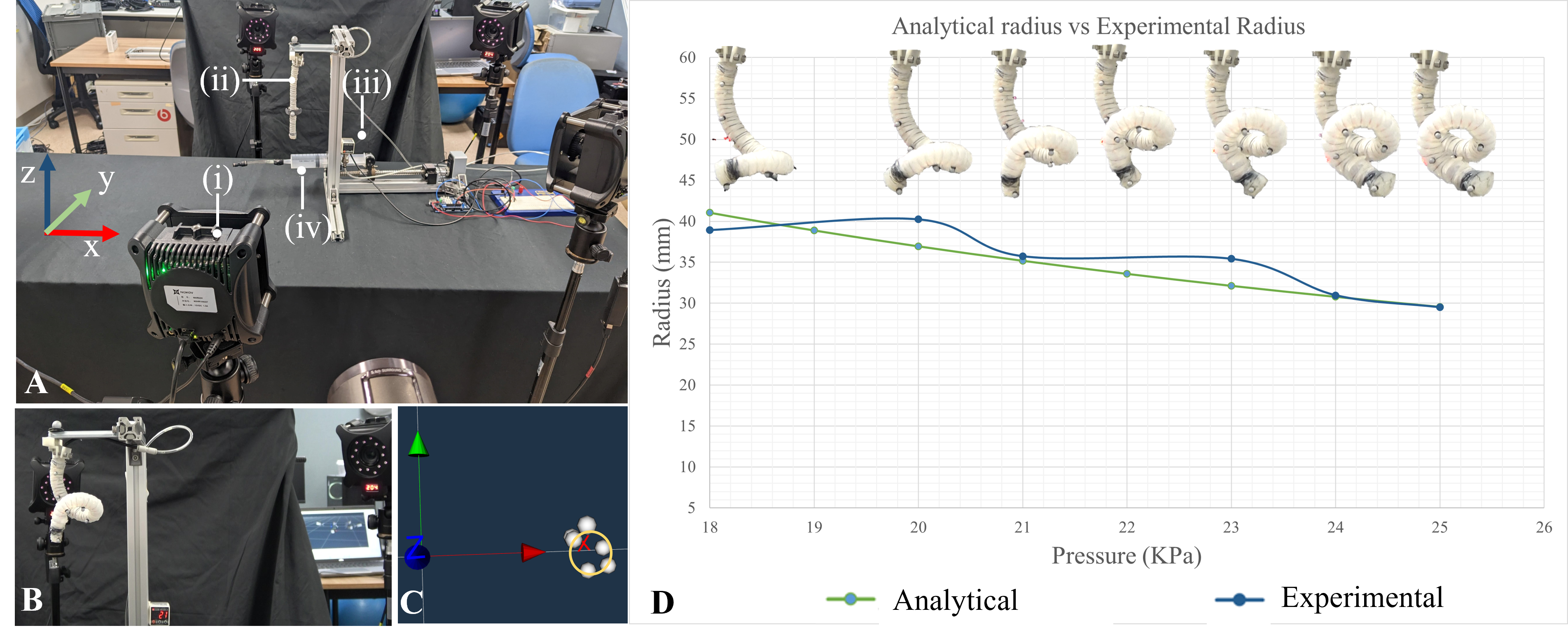}
\caption{Setup and output for the verification of the developed analytical model(A) Motion capture cameras surrounding the actuator setup. (i)Motion capture camera. (ii) Soft actuator. (iii) Pressure sensor. (iv)Air pressure input. (B) Twisting motion with markers on. (C) Fitted circle over markers captured by the
mocap system. (D) Analytical radius vs Experimental radius.
}\label{FigureLabel7}
\end{figure*}%[tbhp]

\subsection{Experimental Verification}
The developed analytical model was verified experimentally. The experimental setup can be seen in Fig. \ref{FigureLabel7}A. Seven markers of 3mm diameter were placed along the edge of the soft actuator(Fig.\ref{FigureLabel7}B). We chose 3mm markers to ensure little to no change in the actuator's motion due to the marker's weight. Smaller markers were tried as well, but the motion capture system failed to detect them in most cases. As for the rest of the setup, six Nokov motion capture cameras were arranged around the actuator. The distance of each camera was set manually to obtain the best output possible. The actuator was actuated with the same pressure input using a ball screw-driven air piston. The Humofit1 was stimulated by pouring hot water over the actuator with a small syringe. This hot stimulation softens the humofit and allows the actuator to execute twisting motion for approximately over a minute. A pressure sensor was used to observe the pressure in the actuator throughout its motion. The actuator was actuated ten times by air pressure input at a slow speed. The markers were observed from the top view, and a circle-fitting method was implemented on them. This was to obtain the radius of the circle which all the marker passes through, i.e., the radius of the twist.

%The actuator was surrounded by six Nokov motion capture cameras, to maximize our chances of recording the whole twisting motion. The actuator was also connected to a pressure sensor. The actuator was equipped with five markers of 3mm diameter along the edge of it. We chose 3mm markers to avoid any deviation in the actuator motion due to the weight of bigger markers. \\

The motion capture provides the coordinates of markers at different pressure values. We observe these markers along the x-y plane and fit a circle through these points(Fig. \ref{FigureLabel7}C). (The topmost marker in this figure is a reference marker placed on the aluminum frame hence it hasn't been included in the circle fitting). The radius of this fitted circle is considered to be the radius of twist $R_t$ as shown in Fig. \ref{FigureLabel6}D. One of the biggest challenges in this verification experiment was not obtaining enough marker points. The 3D nature of twisting motion created occlusion, blocking multiple markers. Hence we actuated multiple times and obtained the experimental radius from the one with the maximum marker points available.   \\

The actuator transitions from a straight line resting state and makes a visible loop state around 18KPa; hence, the model is effective only after this state. This experiment helped us conclude that there is a maximum deviation of 3mm between the twist radius obtained from the analytical model and experimentally as shown in Fig. \ref{FigureLabel7}D. The predicted value gets close to the experimental value as the actuator reaches the final twist state.  This maximum prediction deviation being as small as 10$\%$ is a good improvement from the existing work\cite{Jinag_scaffold}. This proves that the developed analytical model can output Twist radius values with little error and can be used for soft robotics-based precision tasks. 

\section{Soft Robotic Gripper}
A soft robotic gripper was developed by combining three actuators. The multiple motions of each actuator can be used to create multiple motion modes for the gripper. In this section, we share its structure, grasp modes, and our observation of its grasp performance.
\subsection{Structure}
A gripper plate was designed to hold three actuators. This plate was 3D printed with z-ultrat material. The distance between the actuators was decided to be 45mm. This ensured that the tip of actuators would have a few millimetres gap between them, when in an actuated state. This would prevent a failed grasp due to actuators hitting each other. On the other hand, the actuators will be close enough to grasp small-sized objects. Each actuator was given an independent pneumatic supply as different grasp modes require different pressure inputs for each actuator. 

\begin{figure*}[t]
\centering
\includegraphics[width=13.4cm]{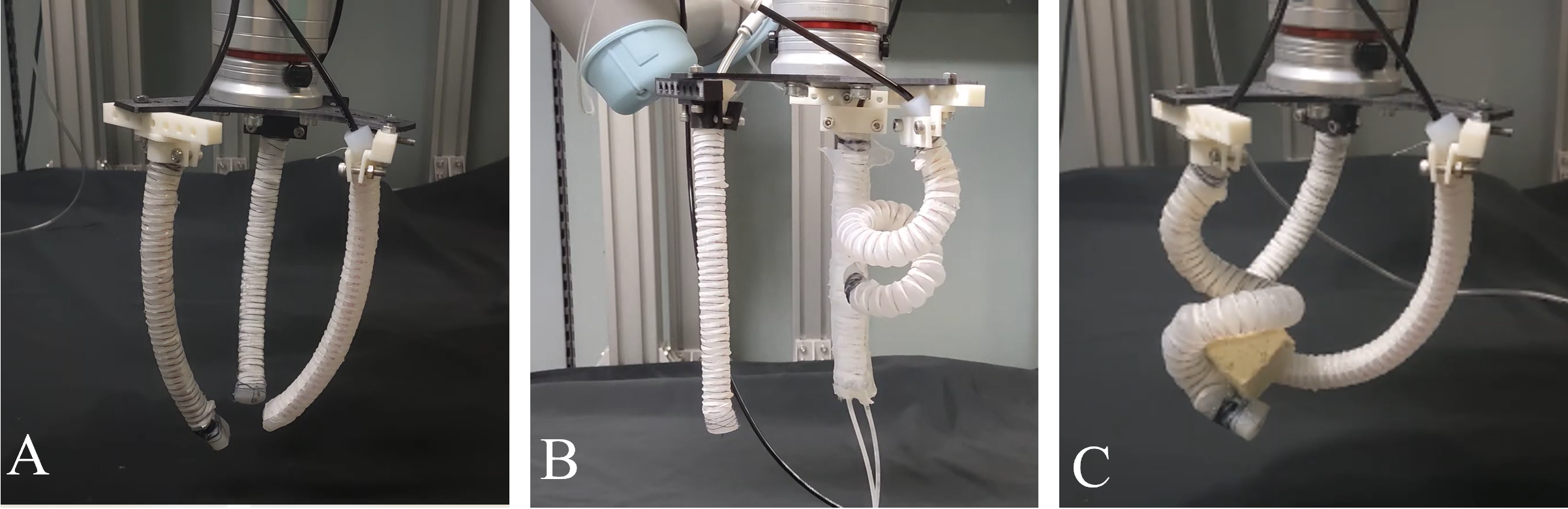}
\caption{Three grasp modes of the gripper (A) Bend mode. (B) Twist mode. (C) Bend-Twist mode.}\label{FigureLabel8}
\end{figure*}%[tbhp]

\subsection{Grasp modes}
The gripper can generate three grasp modes: bending grasp mode, Twisting grasp mode, and Bend-twist grasp mode.  \\

In bending mode, each actuator in the gripper performs a bending motion to achieve grasping. (Fig.\ref{FigureLabel8}A) The bending angle can be adjusted based on the size of the object. Additionally, the gripper can execute various bending radii at the same pressure input(Fig.\ref{FigureLabel5}A). For instance, a small bending radius can be used for an envelope grasp to wrap around larger objects, while a larger bending radius is ideal for creating a pinch grasp for smaller objects. This ability to achieve multiple bending variations significantly expands the gripper’s range of grasping capabilities, even within the bending mode alone. This mode is particularly suitable for grasping standard-shaped objects. To execute this mode, the required pressure input corresponding to the desired bending radius should be applied. As bending motion is the default actuation mode, the bend grasp mode does not require additional stimuli.

In twisting mode, only one actuator from the gripper wraps itself around the target object and forms a helical coil. This grasp ensures a large contact area which helps in grasping delicate and fragile objects. This mode can grasp nonstandard shapes as well due to the adaptability of its twisting nature. This mode can be used to grasp multiple objects too as a single twist actuator can grasp an object by itself.
%(During the grasping process, the gripper makes a pre-grasp pose of radius R. This radius is provided by the system which again depends on the object information. Once the object is centred within the pre-grasp shape, the actuator closes in more and wraps itself around the object.) 
To execute this grasp mode, Humofit1 of one actuator is stimulated with hot water. This results in a singular actuator exhibiting twist as shown in Fig. \ref{FigureLabel8}B.
%To generate the twist bend mode, hot water is passed over the top part of the actuator \ref{FigureLabel4}. The gripper can execute this mode for over a minute without any further stimulus.

The final mode is the combination of both bend and twist modes. In this mode, one actuator coils around a part of the object while the other two actuators bend in and support the object. The bending support is a straightforward bending motion until the actuator makes contact with the object. This method is the most suitable for grasping non-standard objects such as long and thin items, arbitrarily shaped items, etc.
To execute this grasp mode, Humofit1 of one actuator is stimulated with hot water while other fingers need no stimulus. This results in one twist and two bend combination as shown in Fig. \ref{FigureLabel8}C.

Currently, the decision of choosing and executing these grasp modes is done manually but we plan to automate this in our future work by recognizing the object through the system, understanding its parameters such as size, shape, stiffness, etc, and then letting our pre-trained logic choose the most suitable grasp mode.

\subsection{Grasp Performance}
\begin{figure*}[t]
\centering
\includegraphics[width=13.6cm]{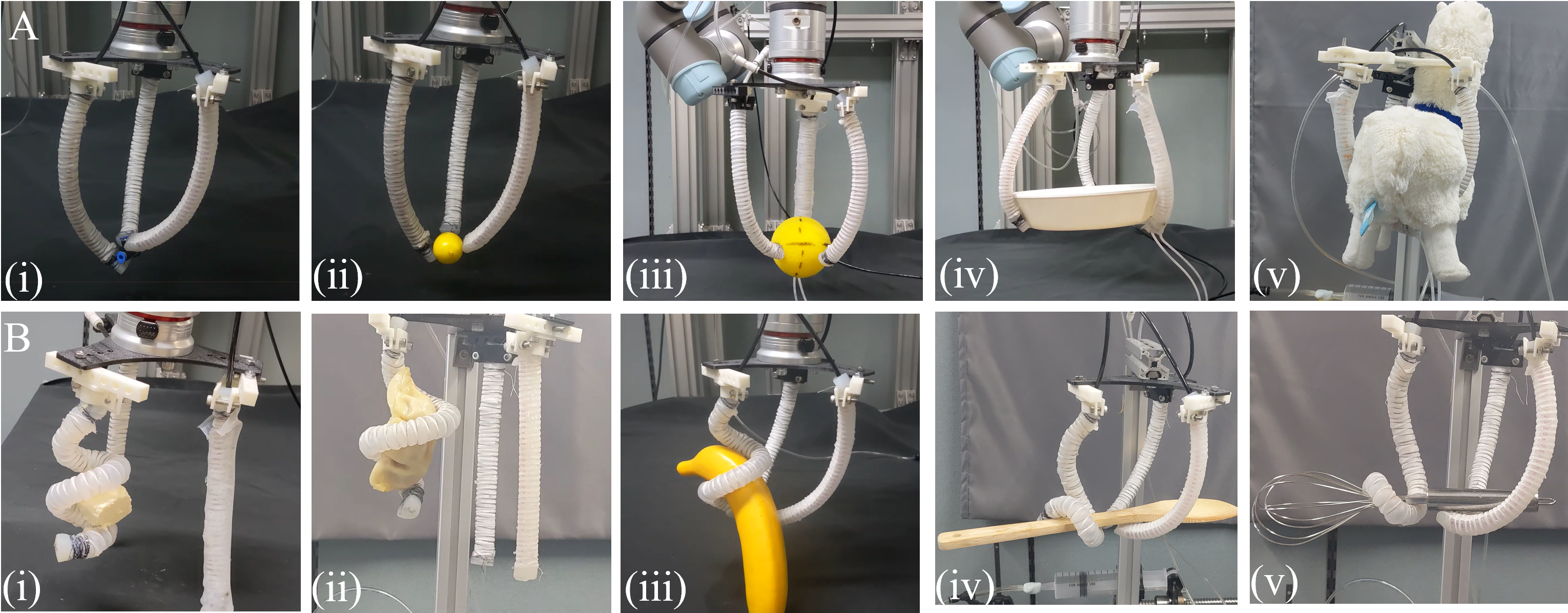}
\caption{Grasping performance on objects of various size, shape, and stiffness (A) (i)10mm valve. (ii) 25mm Berry. (iii) 60mm ball. (iv) 200mm Plate. (v) 300mm Stuffed toy. (B)(i) 30mm
wide Tofu. (ii) 40mm diameter Banana. (iii)40mm Gyoza. (iv) 20m wide Ladle. (v) 25mm long Whisker.
}\label{FigureLabel9}
\end{figure*}

To evaluate the grasping ability and range of the gripper, a grasping experiment was conducted. The developed gripper was attached to a UR-3 arm and objects of various sizes, shapes, and stiffness were grasped. We targeted objects of three categories: Standard shaped objects of different sizes, delicate and soft objects, and non-standard shapes such as long thin ladles etc.   We started with standard-shaped objects of different sizes. The bending mode is chosen as the suitable mode for this category. The gripper succeeded in grasping a 10mm pneumatic valve and a 300mm toy showcasing its range in bending mode as shown in Fig. \ref{FigureLabel9}A(i-v). \\
To represent soft, delicate objects, we chose a cube of Tofu. The actuator gently wraps itself around the tofu creating a large contact area. This grasping pose along with the soft nature of the actuator allows the twist mode to be effective in grasping soft delicate objects with high force and without any damage to the object as shown in Fig. \ref{FigureLabel9}B(i, ii). \\
The third category of objects is a collection of uneven shapes and non-symmetric shapes. For example a banana due to its curved nature. A thin wide ladle or batter mixer. The non-standard, uneven shapes are grasped effectively with a combination of twist and bend modes. One actuator wraps itself or loops around the object while the other two provide support for a stable grasp. Just using twist mode will result in failure due to the long length or weight imbalance of these objects. This mode demonstrated a successful grasp of the mentioned category of objects as shown in Figure \ref{FigureLabel9}B(iii-v).

%[t]

\section{Conclusion}
In this work, we successfully developed a soft robotic actuator that can produce three distinctive motions. Bending, twisting, and extension. This actuator overcame the single-motion mode limitation posed by the existing soft actuators while using a single-air chamber per actuator and only soft materials. We used the structure of a Fiber reinforced actuator and Humofit, a temperature-dependent variable stiffness material, was used as fibers. Further, the actuator’s characteristics were evaluated. It was observed that the actuator has a repeatability of 1mm in bending mode and 5 mm in the case of twisting mode. By changing the temperature of Humofit2, we can change the bending radius of the actuator which allows it to have a 50 percent more sweeping volume than a standard bending actuator. This ability of variable bending radius along with the twisting motion results in a significant increase in the actuator’s range of motion. An analytical model was built to relate the input pressure and the output twist radius for a fiber-reinforced actuator. The accuracy of this model was verified experimentally with the use of a motion capture system. By using three actuators together, we made a robotic gripper that could produce different grasp modes. Each grasp mode presents its advantages which help the gripper in grasping objects of a wide range of size, shape, and stiffness.
Despite the effective performance, our work still has a few limitations. Currently, we use hot or cold water and pass it over the humofit elements via water channels, but this process could be inconvenient to use for a long operation. This limitation can be minimized by making more compact water channels  around the Humofit so that the same performance could be achieved with a smaller amount of water flow. Also, using a closed loop cycle for temperature of the water and Humofit’s temperature and controlling the process accordingly would be some of the examples to fix this limitation. In this study, we've employed the actuator as a soft gripper. Its size can be adjusted to fit various applications, except those demanding an extremely small actuator. As we try to increase motion modes, we would need more separate Humofit elements, and this would increase the chances of temperature crosstalk in small sized actuators. We would explore feasible methods to add more motion modes at extreme sizes in our future research. Future studies will be done to explore the payload capacities and automate the overall grasping task as well. This study can serve as a reference for subsequent research on creating soft actuators with multiple switchable actuation modes.

%\appendix
%\section{Example Appendix Section}
%\label{app1}


%%       \cite{<label>} ==> [1]

%%
%Example citation, See \cite{lamport94}.

  \bibliographystyle{elsarticle-num} 
  \bibliography{references}




%% For numbered reference style

%\bibitem{lamport94}
 % Leslie Lamport,
 % \textit{\LaTeX: a document preparation system},
 % Addison Wesley, Massachusetts,
 % 2nd edition,
  %1994.

%\end{thebibliography}
\end{document}